\documentclass[letterpaper, 10 pt, conference]{ieeeconf}  

\IEEEoverridecommandlockouts                     
\usepackage{color}
\usepackage{xcolor}
\usepackage[]{graphicx}  
\graphicspath{{imgs/}}
\usepackage{booktabs}
\usepackage[caption=false, font=footnotesize]{subfig}
\usepackage{siunitx}

\usepackage{url}
\usepackage{breakurl}
\usepackage[breaklinks]{}

\usepackage{makecell}
\usepackage{amsmath} 
\usepackage{amssymb}  
\usepackage[linesnumbered,ruled]{algorithm2e}

\newcommand{\xhdr}[1]{\vspace{6pt} \noindent {\textbf{#1} }}

\newcommand{\RNum}[1]{\uppercase\expandafter{\romannumeral #1\relax}}

\usepackage{etoolbox}
\makeatletter
\patchcmd{\@makecaption}
  {\scshape}
  {}
  {}
  {}
\makeatother

\title{\LARGE \bf
Realtime Rooftop Landing Site Identification and Selection \\ in Urban City Simulation*
}

\author{Jeremy Castagno$^{1}$, Yu Yao$^{1}$ and Ella Atkins$^{2}$
\thanks{*This work was supported in part by NSF I/UCRC Award 1738714.}
\thanks{$^{1}$Jeremy Castagno and Yu Yao are Robotics Institute PhD candidates, University of Michigan {\tt\small jdcasta@umich.edu}, {\tt\small brianyao@umich.edu}}%
\thanks{$^{2}$Ella Atkins is a Professor of Aerospace Engineering and Robotics, University of Michigan
        {\tt\small ematkins@umich.edu}}%
}

\begin{document}

\maketitle
\thispagestyle{empty}
\pagestyle{empty}

\begin{abstract}

Safe autonomous landing in urban cities is a necessity for the growing Unmanned Aircraft Systems (UAS) industry.  In urgent situations, building rooftops, particularly flat rooftops, can provide local safe landing zones for small UAS.  This paper investigates the real-time identification and selection of safe landing zones on rooftops based on LiDAR and camera sensor feedback.  A visual high fidelity simulated city is constructed in the Unreal game engine, with particular attention paid to accurately generating rooftops and the common obstructions found thereon, e.g., ac units, water towers, air vents. AirSim, a robotic simulator plugin for Unreal, offers drone simulation and control and is capable of outputting video and LiDAR sensor data streams from the simulated Unreal world. A neural network is trained on randomized simulated cities to provide a pixel classification model. A novel algorithm is presented which finds the optimum obstacle-free landing position on nearby rooftops by fusing LiDAR and vision data.

\end{abstract}

\section{Introduction}

Recent advances in Unmanned Aircraft Systems (UAS) perception systems are beginning to enable safe three-dimensional navigation through uncertain environments. Numerous UAS or drone applications include aerial photography, search and rescue, package delivery, and surveillance will benefit. UAS flight operations in densely-populated areas are envisioned to occur above buildings and over people. Safe autonomous landing in urban cities is a necessity, and most prior related research is focused on terrain-based landing \cite{patterson2014timely, atkins2006emergency, di2017evaluating, 6564710}. In urgent situations, building rooftops, particularly flat rooftops, can offer nearby safe landing zones for UAS \cite{desaraju2014vision, Castagno2018ComprehensiveRP}. Urban roofs often have desirable flat-like characteristics and are usually free from human presence \cite{s18113960}. However, landing on urban buildings provides unique challenges such as avoiding the multitude of auxiliary structures scattered across the rooftop.  A database of flat rooftops can be computed and stored \emph{a priori} \cite{s18113960}. Real-time confirmation that the rooftop is unoccupied and approach planning to the landing site are also necessary. 

\begin{figure}[!h]
    \centering
    \includegraphics[width=0.99\columnwidth]{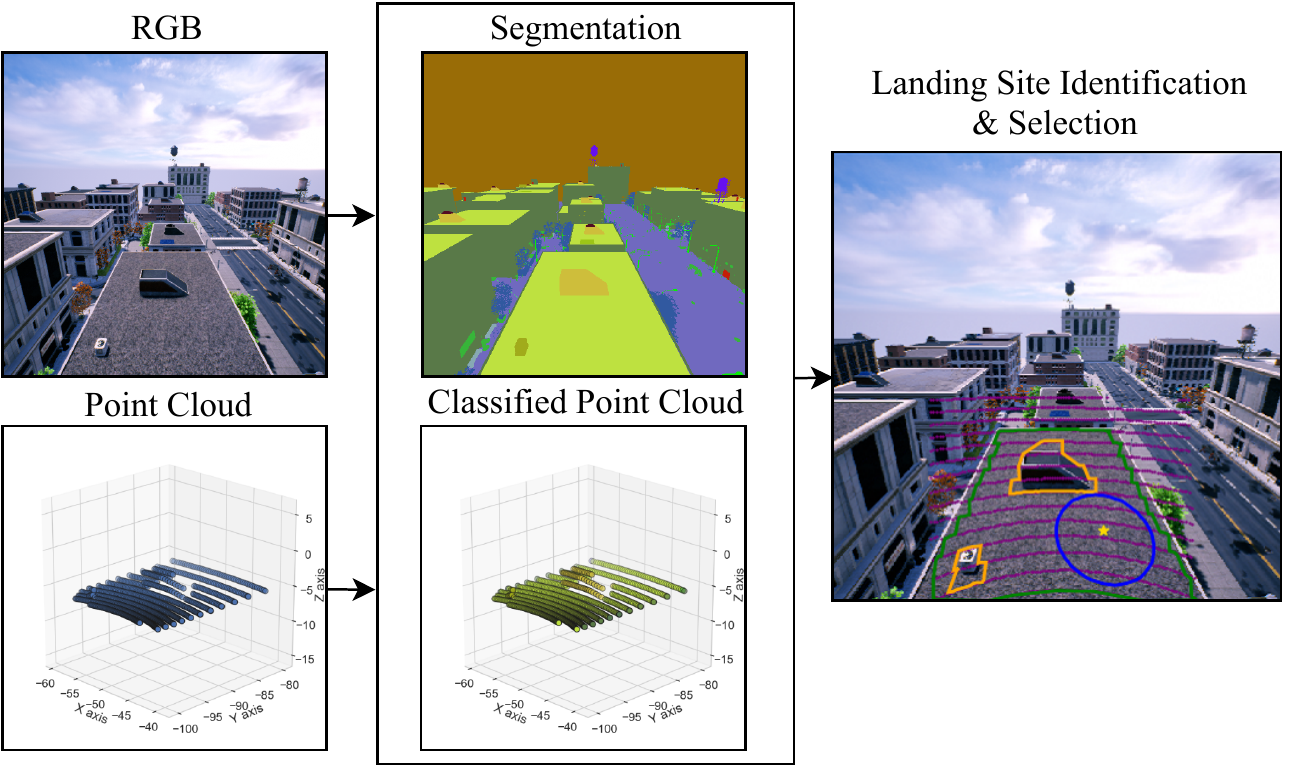}
    \caption{Overview of our landing site selection algorithm. Camera RGB images are transformed into a segmented image on which a point cloud is projected  for point classification. Planar meshes are extracted from the planar point cloud per the right image indicating in green a candidate landing site polygon with orange ``obstacle cutouts''. The blue circle represents the optimal landing site (to 0.5 meter precision) that is flat and obstacle free.}
    \label{fig:ls_overview}
\end{figure}

This paper proposes algorithms to identify and select safe rooftop landing zones in real time using LiDAR and camera sensors.  A high-fidelity simulated city is constructed in the Unreal game engine \cite{unrealengine} with particular attention given to accurate representation of rooftops and common obstacles found thereon, e.g., ac-units, water towers, air vents. AirSim \cite{airsim2017fsr}, a robotic simulator plugin for Unreal, provides simulated drones with video and LiDAR sensors offering hardware in the loop (HIL) sensor feeds of the Unreal simulated environment. A novel landing site selection algorithm is presented that fuses image and LiDAR data to provide the optimal obstacle landing zone on a rooftop within the sensor field of view. Figure \ref{fig:ls_overview} provides a graphical overview of processing steps. Key contributions include:

\begin{itemize}
  \item Construction of a high-fidelity visual city model from real world data of rooftops in Manhattan, New York.
  \item A novel landing site identification and selection algorithm which guarantees finding the optimal landing site from a classified point cloud.  
  \item A comparative study of several state-of-the-art semantic segmentation models to show their classification accuracy as well as time complexity.
  \item Hardware In the Loop (HIL) results of our landing site selection algorithm on desktop and embedded (Jetson TX2) systems with real time performance comparison.
\end{itemize}

\section{Related Work}

Vertical Take-off and Landing (VTOL) aircraft landing and image pixel classification background is summarized below. 

\subsection{VTOL Aircraft Landing Site Selection}\label{section:related_landing}

VTOL aircraft have been flown extensively in unmapped environments. Per  \cite{jin2016board} and \cite{ desaraju2014vision}, terrain landings are commonly investigated, but rooftop landing research is sparse.  In all cases, landing can be decomposed into three steps: landing site identification and selection, trajectory generation, and landing execution \cite{10.1007/978-3-319-16178-5_14}. This paper is focused on rooftop-based landing site identification and selection.

Most research has proposed the use of monocular vision as the primary sensor to aid in landing site identification \cite{jin2016board}.  Single cameras are often lighter, have lower power requirements, are more common,  and are inexpensive. Researchers often use predefined landing site markers and perform image feature matching to recognize the site \cite{yang2015precise, Yang2014}. Our work is focused on landing sites where markers will not be available so that identification must be performed from natural environment features. 

Ref. \cite{desaraju2014vision} specifically identifies candidate landing sites on rooftops.  Using a single camera the authors perform 3D scene reconstruction using structure through motion to generate a disparity map of a rooftop.  They note that the variance of the disparity map along the gravity vector corresponds to the planarity of the landing surface; smaller changes in variance correspond to flatter surfaces. With this assumption the authors apply a kernel filter across the disparity image to identify pixels that are deemed planar. They normalize the resulting image between [0,1] and perform Gaussian process smoothing. This algorithm is run over a downsampled image space and selects the candidate pixel having the ``flattest'' region. Note that this procedure for candidate landing site selection does not consider distance from obstacles. 

Ref. \cite{Forster2015} identifies terrain-based candidate landing sites in an image plane from 2D probabilistic elevation maps generated over terrain. As in \cite{desaraju2014vision}, a monocular camera using structure through motion provides depth information for each pixel. A height discrepancy filter is applied to the depth image to determine planarity, and a distance transform is applied to the image to select the flat pixel farthest away from any non-flat site (pixel). The computational complexity of the algorithm necessitates limiting the size of the map to 100X100 pixels at all altitudes.

An alternative to operating over a discretized image space is modelling the 3D mesh of the roof. The photogrammetry community has recently made significant advancements in building modelling \cite{2016ISPAr49B3, s17092153}. Such techniques provide detailed building surface models but are computationally expensive to compute.  Ref. \cite{doi:10.1080/01431161.2017.1302112} outlines an algorithm to identify planar regions of a building given a point cloud. Our work refines this algorithm in several ways:  separating planar patches if the distance between them is too great, operating on classified point clouds, and adjusting results to sensor noise.  Our algorithm then rapidly converts 3D plane patches into 2D polygons representing flat areas of rooftops.

Selecting the landing zone point furthest from any obstacle is similar to the poles of inaccessibility (PIA) problem studied in \cite{garcia2007poles}. This problem requires finding the point within a 2D polygon that is furthest away from any border. Ref. \cite{polylabel} refines the work in \cite{garcia2007poles} to guarantee global optimality to a specified precision. Key contributions in \cite{polylabel} are the use of quadtrees to divide the search space and a heuristic that efficiently prunes non-optimal landing sites. Our definition of optimality inherits from this work and guarantees a selected flat landing zone is the furthest point away from any obstacle to a specified precision. Note that obstacles in landing zone are represented by holes/cutouts in the polygon and are accounted for. Our work effectively combines these techniques to rapidly identify flat landing zones to determine the optimal landing site location.

\subsection{Semantic Segmentation}\label{section:related_pixel}
Semantic segmentation describes the process of associating each pixel of an image with a class label, such as \emph{sky} or \emph{rooftop}.  Fully convolutional networks (FCN) were first proposed for image semantic segmentation \cite{long2015fully} to learn an end-to-end encoder-decoder model capable of segmentation. The encoder model is a deep CNN that extracts image features with multiple resolutions while the decoder model contains transposed convolutions (upsampling) to predict segmentations with different resolutions. U-Net \cite{ronneberger2015u} further takes advantage of high-resolution features by decoding after each encoding CNN block. SegNet \cite{badrinarayanan2017segnet} is an encoder-decoder model that upsamples from a feature map by storing maxpooling indices from the corresponding encoder layer. Bayesian SegNet \cite{kendall2015bayesian} improves the model by adding dropout layers to incorporate uncertainties in prediction.

Other semantic segmentation work utilizes context-aware models such as DeepLab \cite{chen2017rethinking,chen2018deeplab} and temporal models \cite{siam2018comparative}. Such models have relatively high weights for mobile device applications compared to FCN-based methods. 
This paper comparatively studies the performance of different combinations of lightweight CNN encoders and FCN-based decoders on urban rooftop image semantic segmentation. 

\section{Preliminaries}\label{sec:preliminaries}

We refer to a point cloud as a set of three dimensional points in a local Cartesian reference frame. Each point is defined by the orthogonal bases $\hat{\mathbf{e}}_x$, $\hat{\mathbf{e}}_y$, and $\hat{\mathbf{e}}_z$ where
\begin{equation}
\label{eq:point}
    \vec{{p}_{i}}=x\,\hat{\mathbf{e}}_x+y\, \hat{\mathbf{e}}_y+z\, \hat{\mathbf{e}}_z = [x,y,z]
\end{equation}
A point cloud with $n$ points is represented as set $\mathcal{P} = \{ \vec{{p}_{1}}, \ldots, \vec{{p}_{n}} \}$ where $\vec{{p}_{i}} \in \mathbb{R}^3$. A \emph{classified} point cloud contains an accompanying fourth entry for each point represented as $\mathcal{C} = \{ c_i, \ldots, c_n \}$ where $c_i \in \mathbb{Z}$ denotes a grouping. 

A \emph{plane} in 3D Euclidean space is defined as:
\begin{equation}
\label{eq:plane}
    ax + by + cz + d = 0
\end{equation}
with normal vector $\vec{n} = [a,b,c]$ and distance $d$ from the origin. A rooftop has flat planar patches with \emph{boundaries} but cannot be solely defined as an infinite plane because \emph{spatial connectivity} is required. For example, two flat planar patches may have the same planar equation per Eq. \ref{eq:plane} but be separated by a vertical wall. A method to individually identify such planar patches is outlined in Section \ref{sec:polylidar}.

\section{Landing Site Identification and Selection}\label{section:rooftop_landing}

Our proposed landing site selection algorithm is general to any flat-like surface and will determine the optimal landing position that is obstacle free. The algorithm requires a point cloud of a candidate landing area, with results enhanced if points are classified to determine eligibility for landing per Section \ref{sec:point_cloud_classifcation}. Our algorithm is decomposed into the following processing steps:

\begin{enumerate}
  \item \texttt{Polylidar} - Planar meshes are extracted from a point cloud and rapidly converted into 2D polygons representing a flat region.
  \item Poles of Inaccessibility (PIA) - \texttt{Polylabel} \cite{polylabel, garcia2007poles} is used to rapidly identify the largest inscribed circle within the polygon. This location is guaranteed flat and the point the furthest from any obstacles.
\end{enumerate}

\subsection{Polylidar}\label{sec:polylidar}

Our novel \texttt{Polylidar} algorithm extracts planar meshes from point clouds which are subsequently converted to 2D polygons \emph{with holes intact} in real time.  The steps of the algorithm are:
\begin{enumerate}
  \item 2D Delaunay Triangulation - The point cloud is projected into a 2D space where 2D triangulation occurs.
  \item Triangle Filtering - Triangles are filtered according to a set of configurable criteria: triangle side length, triangle normal, and triangle class.
  \item Planar Mesh Extraction - Flat planer meshes are extracted from spatially connected triangles.
  \item 2D polygon generation - Each planar mesh is rapidly converted to a 2D polygon with holes preserved.
\end{enumerate}

\begin{figure}[!h] 
    \centering
  \subfloat[]{%
       \includegraphics[clip, trim=1.5cm 0.0cm 0.5cm 1.0cm, width=0.55\linewidth]{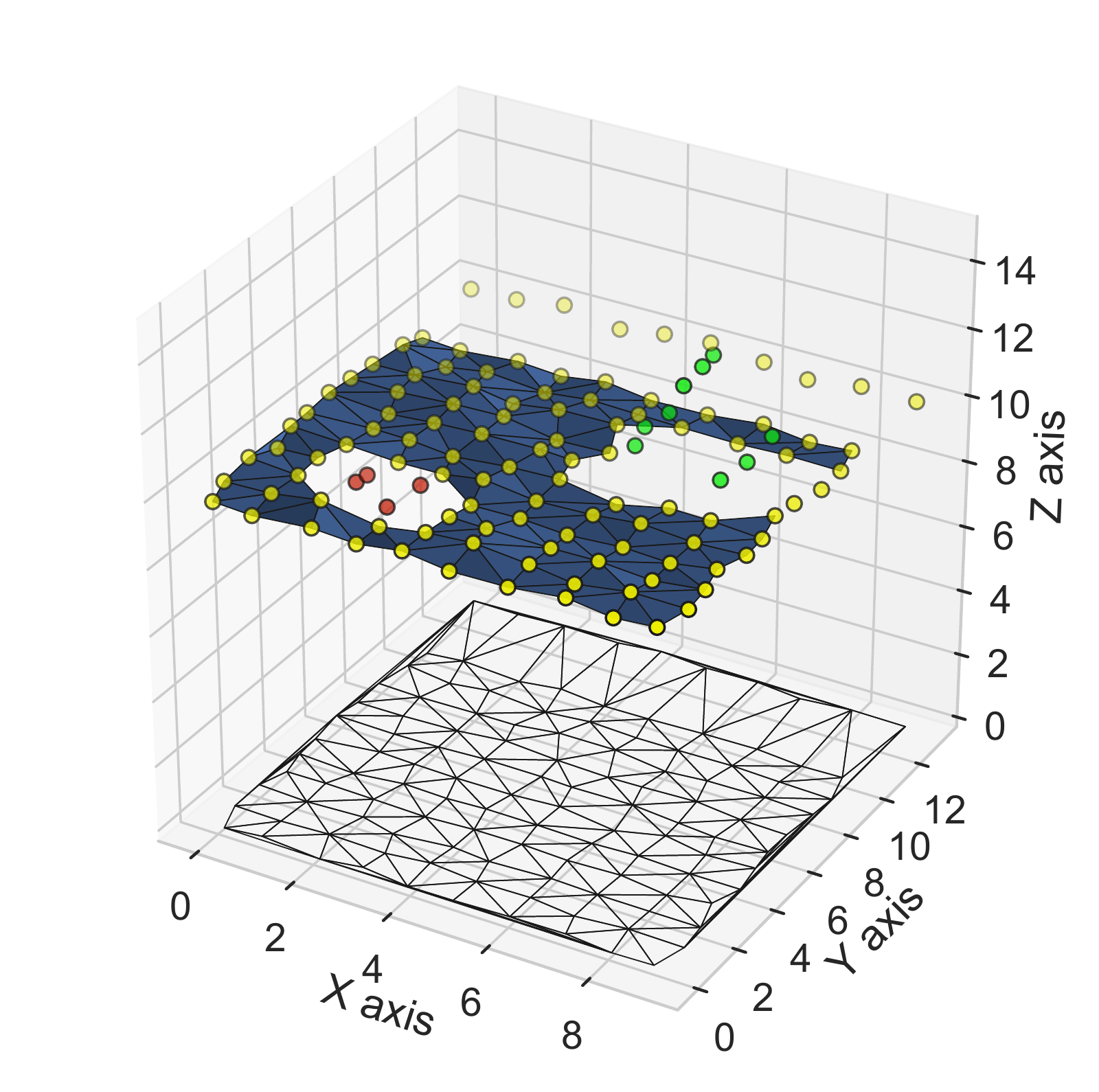}}
    \label{fig:plane_extraction}\hfill
  \subfloat[]{%
        \includegraphics[width=.45\linewidth]{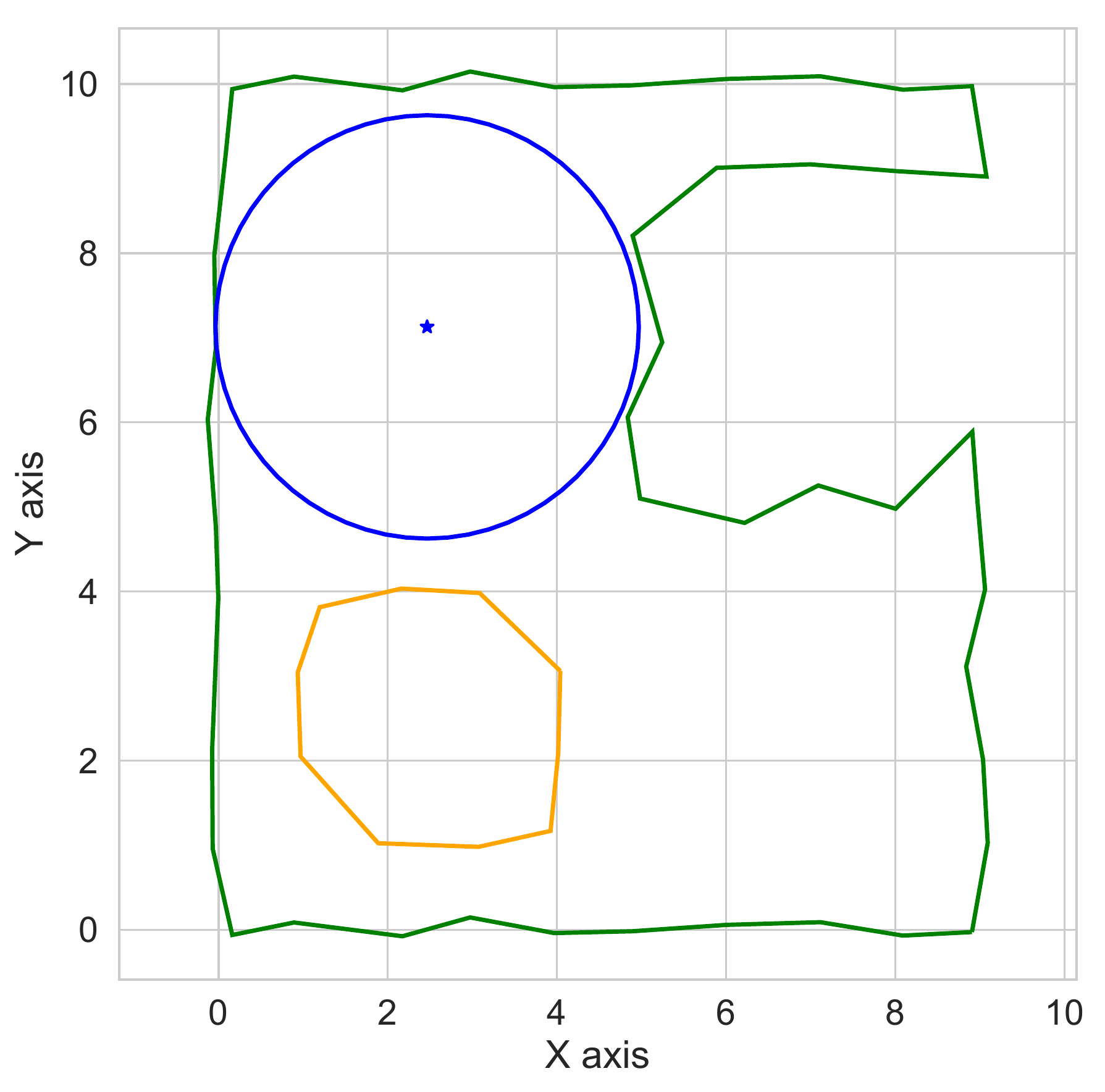}}
    \label{fig:polylabel}\\
  \caption{(a) Demonstration of \texttt{polylidar} extracting planar meshes on a classified point cloud. Red points are from a flat non-rooftop class, yellow points are from the roof class, and bright green points are an unknown obstacle. The point cloud is projected to a 2D space where Delaunay triangulation is performed. Triangles are filtered by size, 3D normals, and by non-roof class. Iterative plane extraction generates the blue flat mesh. (b) \texttt{Polylidar} converts the triangular mesh to a 2D polygon where the green line represents the polygon landing area and orange lines depict holes within the polygon.  \texttt{Polylabel} \cite{polylabel, garcia2007poles} determines the largest inscribed circle within the polygon as depicted in blue.}
  \label{fig:polylidar} 
\end{figure}

An algorithmic description and implementation of Delaunay triangulation can be found in \cite{de2008delaunay,dealaunator, DBLP:journals/corr/Sinclair16a}. The planar mesh extraction step is inspired from \cite{doi:10.1080/01431161.2017.1302112} but modified for use with classified point clouds to add robustness to sensor noise that can lead to spurious ``holes'' in the mesh. The rapid conversion of a flat triangular mesh to a 2D polygon is a novel contribution on its own and is found in our  technical report \cite{jeremytechreport}. Delaunay triangulation returns a set of $k$ triangles defined as
\begin{align}
    \mathcal{T} = \{ t_1, \ldots, t_{k} \}
\end{align}
where each $t_i$ refers to a triangle that holds a set of point \emph{indices} $n_{i,j}$ in  $\mathcal{P}$ defining the vertices of each triangle $t_i$:
\begin{align}
    t_i = \{ \;n_{i,1},\; n_{i,2}, \;n_{i,3}\; \} \\
    n_{i,1},\; n_{i,2}, \;n_{i,3}\in \mathbb{Z}
\end{align}
 Triangle filtering is described in Algorithm \ref{alg:triangle_filtering}. Inputs to the algorithm are the triangles and classified point cloud with parameters set by the user. $l_{max}$ is the maximum length of any side of a triangle, while $dot_{min}$ is the minimum value of the dot product of a triangle normal and the unit $\hat{k}$ vector. $z_{min}$ determines the height threshold of a triangle which if small bypasses normal filtering; this helps remove spurious holes in the mesh due to sensor noise.  $\mathcal{V}$ represents the set of point classes allowed for plane extraction. Filtering a triangle by side length ensures that points are not spatially connected when the distance between them is too large, while normal filtering removes non-planar surfaces.
 
 Planar mesh extraction is shown in Algorithm \ref{alg:all_plane_extraction}. The method begins by picking a random seed triangle from the set of filtered triangles.  This seed triangle is then continually expanded through its adjacent neighbors until all triangles forming its boundary are added to a complete plane. This process is repeated until all filtered triangles have been expanded. A set of triangle mesh planes are returned, and the largest is used for landing site selection. The triangle mesh is then converted to a 2D polygon with holes intact.

\begin{algorithm}\label{alg:triangle_filtering}
    \SetKwInOut{Input}{Input}
    \SetKwInOut{Output}{Output}
    \SetStartEndCondition{ }{}{}%
    \SetKwProg{Fn}{def}{\string:}{}
    \SetKwFunction{Range}{range}
    \SetKw{KwTo}{in}\SetKwFor{For}{for}{\string:}{}%
    \SetKwIF{If}{ElseIf}{Else}{if}{:}{elif}{else:}{}%
    \SetKwFor{While}{while}{:}{fintq}%

    \Input{Triangles: $\mathcal{T}$, Point Cloud: $\mathcal{P}$, Point Class: $\mathcal{C}$ \\
           $l_{max}, dot_{min}, z_{min}, \mathcal{V} $}
    \Output{Filtered Triangle Set, $\mathcal{T}_f$}
    
    $\mathcal{T}_f = \emptyset$ \\
    \For{$t_i$ in $\mathcal{T}$ }{
        $\hat{{n}}_{i} = \operatorname{normal}(t_i)$\\
        $dot_i = |\hat{{n}}_{i} \cdot \hat{k}|$\\
        $l_{max,i} = \operatorname{MaxTriangleEdgeLength}(t_i)$ \\
        $z_{max,i} = \operatorname{MaxTriangleVerticalHeight}(t_i)$ \\
        $v_i = 1 \; if \; \forall n_{i,j} \in t_i \;, \mathcal{C}(n_{i,j}) \in \mathcal{V} $\\ 
        \uIf{$l_{max,i} < l_{max} $ {\bf and} $dot_i > dot_{min}$ {\bf and} $v_i = 1$}{
            $\mathcal{T}_f$ = $\mathcal{T}_f$ + $t_i$
        }
        \uElseIf{$l_{max,i} < l_{max} $ {\bf and} $v_i = 1$ {\bf and} $z_{max,i} < z_{min}$} {
            $\mathcal{T}_f$ = $\mathcal{T}_f$ + $t_i$ \tcp*{prevent hole in mesh}
        }
    }
    return $\mathcal{T}_f$
    \caption{Triangle Filtering}
\end{algorithm}

\begin{algorithm}\label{alg:all_plane_extraction}
    \SetKwInOut{Input}{Input}
    \SetKwInOut{Output}{Output}

    \Input{Filtered Triangle Set, $\mathcal{T}_f$}
    
    \Output{A set of $n$ planes, $\mathcal{L} = \{l_i, \ldots, l_n\}$ \\ 
        where $l_i = \{t_j, \ldots, t_m\}$}
    $\mathcal{L} = \emptyset$ \\
    \While{$\mathcal{T}_f$ is not empty}{
        $t_s= \operatorname{RandomChoice}(\mathcal{T}_f)$  \tcp*{seed triangle}
        $l = \emptyset$ \tcp*{new planar mesh}
        $\mathcal{T}_f = \mathcal{T}_f \setminus t_s $ \\
        Initialize $queue$ with $t_s$\\
        \While{$queue$ is not empty}{
            $t = \operatorname{pop}(queue)$\\
            $l = l + t$\\
            \For{neighbor $\text{adjacent to}$ $t$ and $neighbor \in \mathcal{T}_f$ }{
                $\operatorname{add}(queue, neighbor)$ \\
                $\mathcal{T}_f = \mathcal{T}_f \setminus neighbor $ \\
            }
    
        }
        $\mathcal{L} = \mathcal{L} + l$ 
    }
    return $\mathcal{L}$
    \caption{Planar Mesh Extraction}
\end{algorithm}



Figure \ref{fig:polylidar} demonstrates the algorithm being performed on an example building. Yellow points are from the roof class, red points from a flat \emph{non}-rooftop class, and bright green points are an unknown obstacle. The algorithm begins first by projecting the point cloud into 2D space and performing a Delaunay triangulation as shown in Figure \ref{fig:polylidar}a.  Note that though the triangulation occurs in 2D space, a 3D representation of each triangle can be constructed from corresponding 3D points. Afterwards each triangle is filtered according to Algorithm \ref{alg:triangle_filtering}, while Algorithm \ref{alg:all_plane_extraction} extracts the plane shown in blue. Note that the triangles connected to the red non-roof points have been removed (class filtering) and a hole is observed in the mesh. Green point triangles are non-planar and removed as well via normal filtering.  The points and associated triangles at the far end of the $y$-axis were too far apart per triangle length filtering so the mesh does not extend further back. 

Figure \ref{fig:polylidar}b demonstrates the conversion of the blue mesh to a 2D polygon where the green line represents the flat polygon landing area and the orange line depicts a hole in the polygon. The greatest inscribed circle is calculated for this polygon using \texttt{polylabel} whose circle is shown in blue.  The center of the circle is the optimal position for landing in reference to planarity and distance from obstacles. This radius specifies a distance guaranteed to be obstacle free for the UAS itself or operators to consider.


\section{Simulation Testbed}\label{section:simulation_testbed}

Section \ref{section:rooftop_analysis} provides an overview of our analysis of flat rooftops in Manhattan, New York, while Section \ref{section:rooftop_random} discusses random rooftop generation in the Unreal environment. 

\subsection{Analysis of Rooftops}\label{section:rooftop_analysis}

Before constructing the simulation environment, an analysis of rooftops in Manhattan was performed. Since our work is focused on flat rooftop landing sites, only flat-like roofs in Manhattan were sampled \cite{s18113960}. Data was collected manually by inspecting high resolution satellite and aerial imagery of buildings and recording the rooftop assets and associated quantities observed. A total of 112 buildings were randomly sampled from Manhattan near the Southwest corner of Central Park. 
Table \ref{table:roof_quantity} lists the 12 most common items found on a building rooftop in midtown Manhattan and the average quantity observed. If a building does not contain an asset its quantity is recorded as zero. 

\begin{table}[ht]
\centering
\caption{Common rooftop items with average quantities}
\label{table:roof_quantity}
\begin{tabular}{@{}ll@{}}
\toprule
Item                   & Mean Quantity \\ \midrule
air-vents              & 1.12          \\
small-rooftop-entrance & 0.88          \\
skylight               & 0.51          \\
small-building         & 0.45          \\
ac-unit                & 0.28          \\
seating                & 0.12          \\
air-ducts              & 0.11          \\
water-tower            & 0.10          \\
chimney                & 0.05          \\
enclosed-water-tower   & 0.04          \\
tarp                   & 0.03          \\
vegetation             & 0.02          \\ \bottomrule
\end{tabular}
\end{table}

\begin{figure*}[ht!]
\centering
\includegraphics[width=.99\linewidth]{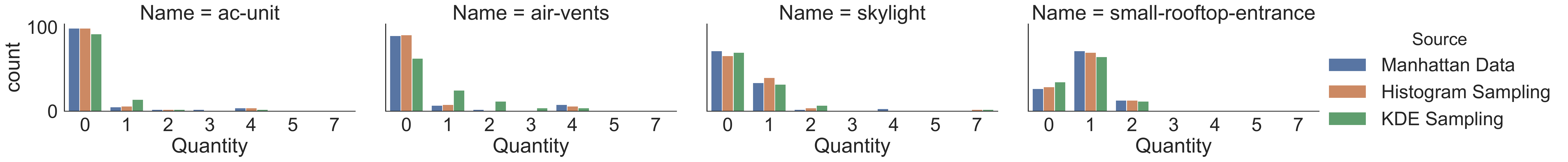}
\caption{Histogram (blue) of four common rooftop items sampled from Manhattan. Two different sampling techniques are shown, directly sampling from a histogram (orange) and sampling from a fitted kernel density estimator (blue).  }
\label{fig:histogram_sampling}
\end{figure*}

We constructing simulated urban cities with rooftop assets following the distributions observed from the Manhattan dataset. Histograms of four common rooftop assets are shown in Figure \ref{fig:histogram_sampling} and marked in blue. Instead of sampling from an assumed distribution and fitting parameters to the data, one can instead directly sample from the histogram distribution providing results shown in orange. Another common method is fitting a kernel density estimator (KDE) to the data and sampling from it as shown in green. Both methods provide nearly identical results except for the air-vents distribution which has large discrete jumps in quantities. The KDE smooths these jumps while sampling from the histogram maintains the distributions observed in the data.

\subsection{Unreal Engine Random World Generation}\label{section:rooftop_random}

Game assets for urban city buildings were purchased from \cite{urbancity} while high quality rooftop assets (e.g., ac units, water towers) were purchased from \cite{urbanrooftop}. Assets were modified to allow configurability in textures and materials to enhance world diversification. For example, an air-vent can be configured to take on a variety of different metal textures and reflectivity properties. A base city was constructed with over 30 buildings having a variety of textures, sizes, and shapes. Once the base city was constructed several random worlds were generated for training and testing purposes. Note that for this work randomness is focused in building rooftop assets, not the buildings themselves. 

A world generation script adds rooftop assets to existing buildings within the base map.  The script takes as input a 2D map of buildings in the environment and a list of possible assets to generate. The script supports configurable options including: probability and quantity of asset placement, spatial location on the rooftop, asset orientation, and appearance properties (e.g. materials/textures). Note that the likelihood of a new asset placement is assumed independent of assets already placed on a building for simplicity. 

Training a neural network to segment and classify rooftop assets requires training data that has numerous labeled examples of all rooftop assets. To provide this training data random worlds were generated where equal probabilities were assigned to asset candidate placements. This leads to diverse yet unrealistic worlds, e.g. more than than the expected number of water towers on a roof. For test sets the worlds were generated according to the quantity distribution of assets sampled from the Manhattan dataset. Direct sampling from the data histogram was performed and leads to a more realistic world in terms of asset quantities, though placement could still be considered odd. Finally a manually-constructed world was generated according to the roof asset distribution in Manhattan but given more careful attention to asset placements in relation to one another.


\section{Point Cloud Semantic Classification}
To classify each point in the LiDAR point cloud, we first apply an image semantic segmentation algorithm to RGB camera images.  Sections \ref{sec:cnn_backbone} and \ref{sec:cnn_opt} compare semantic segmentation models, while Section \ref{sec:point_cloud_classifcation} outlines point cloud classification using a predicted segmented image.


\subsection{Combinations of CNN Backbones and Meta-architectures}\label{sec:cnn_backbone}

We implemented and evaluated different image semantic segmentation models  \cite{siam2018comparative} on two different hardware platforms: a desktop with RTX 2080 and a Jetson TX2.

\xhdr{CNN backbones} are feature extractors or encoder networks that downsample input images to obtain high-dimensional features. This paper  compared two backbone CNN networks: MobileNets~\cite{howard2017mobilenets} and ShuffleNet~\cite{zhang2018shufflenet}. MobileNets are light weight deep neural networks designed for mobile devices. A standard convolution operation is factorized into a depth convolution and a pointwise convolution, termed depthwise separable convolution. ShuffleNet generalizes depthwise separable convolution and group convolution to achieve an efficient CNN encoder for a mobile device. A channel shuffle operation is applied to realize the connectivity between the input and output of different grouped convolutions.

\xhdr{Meta-architectures} are upsampling or decoder networks that reconstruct a segmentation image from downsampled feature maps. This paper compares two meta-architectures based on \cite{siam2018comparative}: FCN~\cite{long2015fully} and U-Net~\cite{ronneberger2015u}. FCN combines CNN features from different depths of the encoder network during upsampling to utilize the information from a higher resolution image. The FCN model applied in this paper combines feature maps from \textit{pool3, pool4} and \textit{conv7} layers to achieve better precision, known in FCN as stride 8 or FCN8s. U-Net takes advantage of the higher resolution feature by upsampling from each stage of the CNN encoder. At the end of each CNN block, the feature map is both input to the next CNN block and combined with the upsampled feature map;  upsampling continues until a final segmentation map is created.

\subsection{Computation graph optimization}\label{sec:cnn_opt}
Although the selected backbone networks are lightweight, we needed to further optimize the computation graph to achieve real-time performance on a mobile device that might be carried onboard a UAS. We first remove all  training nodes since the model will only be used for inference during flight. Next, switch nodes created by conditional operations are removed since the inference condition is always known, and redundant operations such as identity and merge nodes are also removed. Finally, the TensorRT\footnote{https://developer.nvidia.com/tensorrt} graph optimizer is applied to optimize float computations.

\subsection{Point Cloud Classification}\label{sec:point_cloud_classifcation}
A monocular RGB camera provides an $MxN$ image processed through a neural network to construct a segmented image. Every pixel of this image takes an integer value of a predicted class. Each point in the point cloud $\mathcal{P}$ is represented in the local reference frame per Eq. \ref{eq:point} but can be expressed in the camera frame as  $p_i^c = [x^c, y^c, z^c]$. The projection of each point into camera image coordinates is:

\begin{align} 
u &= f_{x} \frac{x_i^{c}}{z_i^{c}}+c_{x} \\ 
v &= f_{y} \frac{y_i^{c}}{z_i^{c}}+c_{y} 
\end{align}
where $u$ and $v$ are image pixels, $f_x$ and $f_y$ are camera focal lengths in pixel units, and $c_x$, $c_y$ are camera principle point offsets. Some projected points may be outside the camera image and are discarded, e.g. $u,v$ $<$ 0.  The integer value of each pixel is recorded and stored in point class data structure $\mathcal{C}$. For this work the primary class of interest is the ``rooftop''.
\begin{table*}[ht!]
    \centering
    \scriptsize
    \caption{The per-class IoU and mean IoU of different image semantic segmentation networks on our urban rooftop dataset. The top two IoUs of each class and the mean IoU are bold. Chimney class is absent in the Random dataset therefore the IoUs are not available.}
    \begin{tabular}{c|c|c|c|c|c|c|c|c|c|c|c|c|c|c|c}
    \toprule
         & sky & ground & \makecell{building\\ wall} & \makecell{building\\ rooftop} & \makecell{small\\ rooftop\\ entrance} & \makecell{sky-\\light} & \makecell{air\\ vents} & \makecell{ac-\\unit} & \makecell{sea-\\ting} &\makecell{air-\\ducts}& \makecell{water\\ tower} & \makecell{chim-\\ney} & tarp & \makecell{vege-\\tation}  &  \makecell{Mean\\ IoU} \\
        \midrule
        \makecell{MobileNet\\ + FCN8s} & 0.99 & \textbf{0.93} & \textbf{0.96} & \textbf{0.98} & \textbf{0.82} & \textbf{0.79} & \textbf{0.48} & \textbf{0.78} & \textbf{0.42} & \textbf{0.80} & 0.74 & N/A & \textbf{0.90} & 0.81 & \textbf{0.74} \\
        \midrule
        \makecell{ShuffleNet\\ + FCN8s} & 0.99 & 0.92 & 0.95 & 0.97 & 0.78 & 0.76 & 0.41 & 0.75 & 0.37 & 0.73 & 0.68 & N/A & 0.84  & 0.79 & 0.71 \\
        \midrule
        \makecell{MobileNet\\ + UNet} & 0.99 & 0.93 & \textbf{0.97} & \textbf{0.98} & \textbf{0.83} & \textbf{0.84} & \textbf{0.50} & \textbf{0.81} & \textbf{0.47} & \textbf{0.81} & \textbf{0.78} & N/A & 0.84 & \textbf{0.90} & \textbf{0.76} \\
        \midrule
        \makecell{ShuffleNet\\ + UNet} & 0.99 & \textbf{0.94} & 0.96 & 0.98 & 0.79 & 0.79 & 0.36 & 0.77 & 0.34 & 0.74 & \textbf{0.76} & N/A & \textbf{0.91} & \textbf{0.88} & 0.73 \\
        \bottomrule
    \end{tabular}
    \label{tab:per_class}
    \vspace{-12pt}
\end{table*}

\section{Experimental Results}

\subsection{Rooftop Image Semantic Segmentation}

\begin{table}[]
    \centering
    \caption{Performance of semantic segmentation models on RTX 2080 and TX2}
    \begin{tabular}{c|c|c|c|c}
    \toprule
         &  \multicolumn{2}{c|}{RTX 2080} & \multicolumn{2}{c}{TX2}  \\
         \cline{2-5}
         & mIoU &  time(ms) & mIoU & time(ms)  \\
        \hline
        \makecell{MobileNet\\ + FCN8s} & 0.74 & 8.17 & 0.74 &  88.49 \\ 
        \makecell{ShuffleNet\\ + FCN8s} & 0.71 & 7.44 & 0.71 & 74.06 \\ 
        \makecell{MobileNet\\ + UNet} & 0.76 & 18.47 & 0.76 & 261.94 \\
        \makecell{ShuffleNet\\ + UNet} & 0.73 & 17.17 & 0.73 & 146.47\\
    \bottomrule
    \end{tabular}
    \label{tab:2080_vs_tx2}
    \vspace{-10pt}
\end{table}

\xhdr{Implementation details:} We evaluated four models for image semantic segmentation: MobileNet+FCN8s, ShuffleNet+FCN8s, MobileNet+UNet and ShuffleNet+UNet. We modified models based on tensorflow implementations \cite{siam2018comparative} and perform training on a system with an Nvidia RTX 2080 GPU. Each model was trained for 100 epoches and the best models were applied in testing. All models are tested on both a RTX 2080 GPU and a Jetson TX2 with Jetpack 3.3 to evaluate real-time on-board performance.

\xhdr{Metrics:} We show the mean intersection over union (IoU) and per-class IoU for each method. 

\begin{figure}[!htb]
    \centering
    \includegraphics[width=1.0\columnwidth]{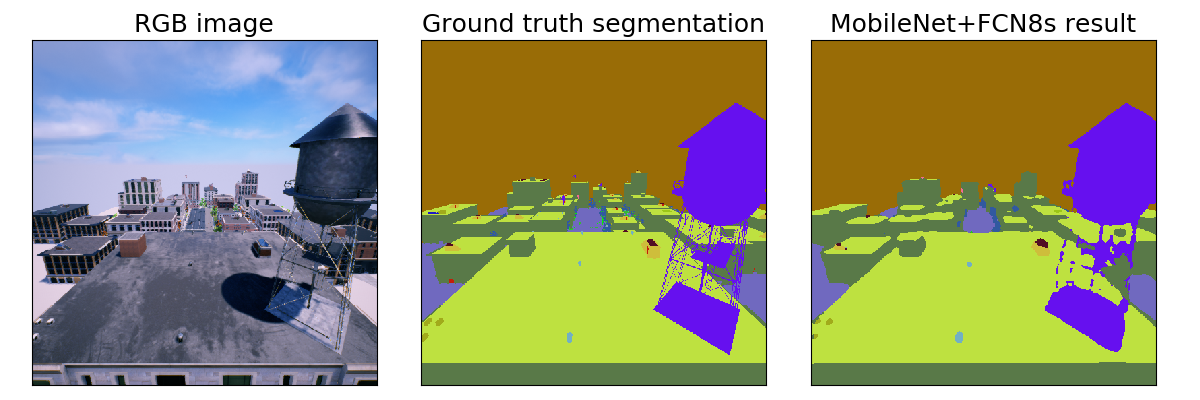}
    \caption{Semantic segmentation on an Unreal rooftop image.}
    \label{fig:seg_example}
    \vspace{-10pt}
\end{figure}

\xhdr{Quantitative results:} As can be seen in Table~\ref{tab:per_class}, the MobileNet+UNet model achieves the best mean IoU and the best per-class IoU on most classes while  MobileNet+FCNs performs the second best in most Table \ref{tab:per_class} cases. Specifically both models outperform the ShuffleNet based methods on small-rooftop-entrance, skylight, air-vents and ac-units which appear frequently in the real world and are more important to rooftop landing tasks. To select the model to be applied in an on-board system, we compare the performance of all four models on both RTX 2080 and TX2 in Table~\ref{tab:2080_vs_tx2}. It can be seen that even with graph optimization applied, all models are around 10 to 15 times slower when running on the TX2. The FCN8s meta-architecture is a lighter model and thus faster than the UNet model. The MobileNet backbone is generally slower than ShuffleNet but achieves better mean IoU. We selected MobileNet+FCN8s as the image semantic segmentation method in the following experiments to achieve a balanced trade-off between accuracy and timing performance. An example RGB image, ground truth label and MobileNet+FCN8s predicted image are presented in Fig.~\ref{fig:seg_example}.

\subsection{Landing Site Identification and Selection}

AirSim simulations were executed on a high end gaming desktop. The landing site algorithm was run on both this desktop and Jetson TX2. The AirSim drone carried a monocular camera and 16-beam scanning LiDAR sensor with vertical and horizontal FOVs of $40^{\circ}$ and $60^{\circ}$, respectively. The LiDAR sensor rotated at 10 Hz and collected 100,000 pt/s. For simplicity, only one rotational scan of the LiDAR sensor was used to collect a point cloud of the landing site surface as input to our algorithm. The following algorithm parameters chosen were: $l_{max}=\SI{4.00}{\meter}, dot_{min} = 0.96, z_{min} = \SI{0.10}{\meter}, \mathcal{V} = \{\;rooftop \;\}$. The \texttt{polylidar} and \texttt{polylabel} algorithms were written in C++, while point classification ran in Python using the library NumPy. The average point cloud size was $\sim 1500$. Note that the landing site algorithm is limited by Delaunay triangulation (with respect to point cloud size) which scales with $O(n\log{}n)$\cite{dealaunator}.

\begin{figure}[!htb] 
    \centering
  \subfloat[]{%
       \includegraphics[width=0.49\linewidth]{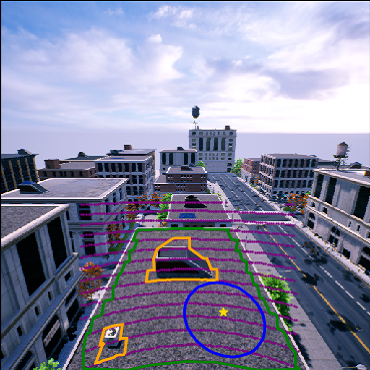}}
    \label{fig:landing_obstacle}\hfill
  \subfloat[]{%
        \includegraphics[width=.49\linewidth]{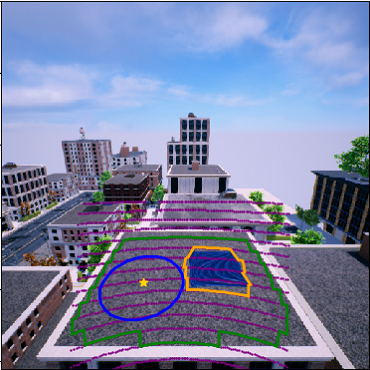}}
    \label{fig:landing_tarp}\\
  \caption{Sample results of landing site selection. All colored overlays are points and planes in 3D space projected into the image. The purple dots represent the point cloud, while the extracted flat plane is denoted by the green polygon and orange holes. The chosen landing site and safe radius are depicted by a star and blue circle, respectively. (a) Detected building rooftop-entrance and ac-unit. (b) Detected flat-like blue tarp.}
  \label{fig:example_landing_detection} 
\end{figure}

Figure \ref{fig:example_landing_detection} shows two example scenarios of a landing site being detected from the perspective of the drone. All colored overlays are points and planes in 3D space projected into the image. The purple dots represent the point cloud, the green outline and orange lines represent the extracted flat plane boundary and holes, respectively. The blue circle is the greatest inscribed circle in the extracted plane with the star marking the safest landing position. Figure \ref{fig:example_landing_detection}a shows two obstacles, a rooftop entrance and an ac-unit, being detected and removed from the green landing zone encompassing the roof.  Likewise Figure \ref{fig:example_landing_detection}b shows a flat-like blue tarp being detected as an obstacle (by point class) which otherwise would have been missed if using depth information only. This scenario demonstrates the enhanced security provided through multi-modal data fusion.

\begin{figure}[!htb]
\centering
\includegraphics[width=.90\linewidth]{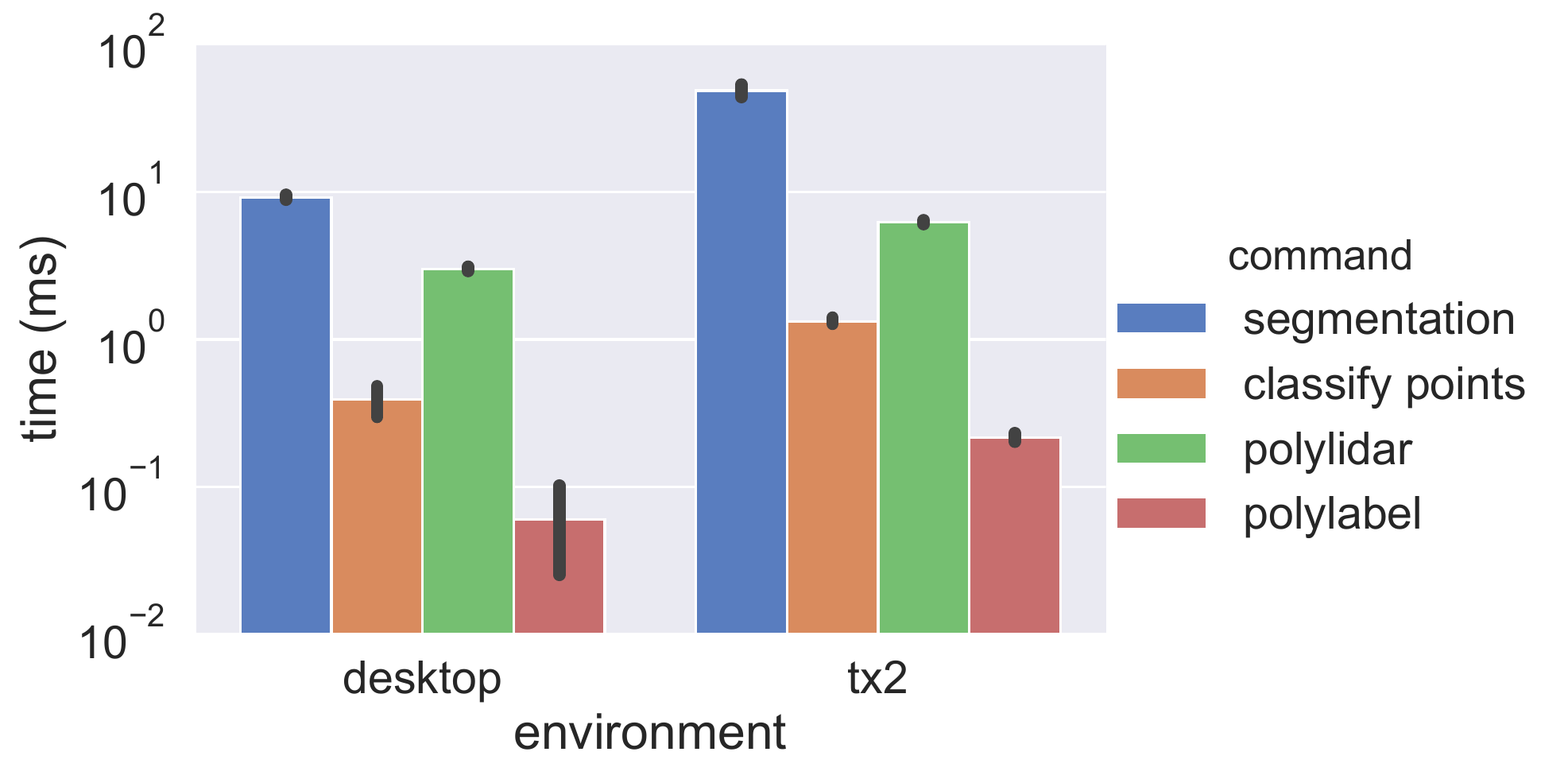}
\caption{Execution time results for landing site selection on a desktop (RTX 2080) and Jetson TX2. Time is denoted with a logarithmic scale. }
\label{fig:landing_time}
\end{figure}

Execution times for each algorithm step were recorded for over 130 landing scenarios from 30 separate buildings. Figure \ref{fig:landing_time} displays these times with 95\% confidence intervals for both the desktop system and embedded TX2 board. As expected the desktop implementation executes much faster, especially the GPU segmentation step. However, all other steps only doubled in time when executed on the TX2. Note that segmentation and point classification are only required to enhance our landing site selection algorithm. On the TX2 \texttt{polylidar} and \texttt{polylabel} can execute in $\sim\SI{7}{ms}$ on an unclassified point cloud missing only flat-like obstacles observable through a camera sensor (e.g., tarps, boards).

\section{Conclusions and Future Work}

This paper has presented a real time landing site identification and selection algorithm that finds the optimal (largest and obstacle free) flat landing surface on a building rooftop within sensor field of view.  The algorithm was evaluated in a high fidelity simulated city in Unreal Engine based on real world rooftop data from midtown Manhattan, New York. Multiple semantic segmentation neural networks were trained and evaluated with MobileNet + FCN8s chosen for its high accuracy and speed. By fusing camera and LiDAR sensor data into a classified point cloud our algorithm identifies obstacles that otherwise would have been missed with only depth information.

Though our simulated city models rooftop obstacles accurately, future work is required to build rooftop models with a higher level of fidelity to cities such as Manhattan (building height, textures, etc.). Additionally, our current simulation environment only generates sunny weather image data thus requires extension to more general lighting and weather conditions. Currently an obstacle may be observed in the camera but missing in the point cloud (points are absent); our landing site selection algorithm will not account for such obstacles. Future work will investigate how to further prune \texttt{polylidar} extracted mesh results with vision data. Real-world experiments of our landing site identification and selection algorithm will also be performed in the future.

\addtolength{\textheight}{-12cm}   


\bibliographystyle{unsrt}
\bibliography{reference}

\end{document}